\documentclass[conference]{IEEEtran}
\usepackage{graphicx}
\usepackage{amsmath}
\usepackage{amsfonts}
\usepackage{cite}
\usepackage{url}
\usepackage{booktabs}
\usepackage{multirow}
\usepackage{array}
\usepackage{caption}
\usepackage{algorithm}
\usepackage{algorithmic}

\begin{document}

\title{WISE-HAR: A Generalizable Ensemble Deep Learning Framework for WiFi-Based Human Activity Recognition}

\author{\IEEEauthorblockN{Maheen Arshad, Qindeel E Zahra, Muhammad Khuram Shahzad}
\IEEEauthorblockA{Department of Computing, School of Electrical Engineering and Computer Science\\
National University of Sciences and Technology (NUST), Islamabad, Pakistan\\
Email: {marshad.msit25seecs, qzahra.msit25seecs, mkhuram.shahzad}@seecs.edu.pk}
}

\maketitle

\begin{abstract}
Human Activity Recognition (HAR) using WiFi signals has emerged as a transformative technology for smart homes, healthcare monitoring, security systems, and ambient assisted living. Unlike traditional camera-based systems that raise significant privacy concerns and fail in low-light conditions, or wearable sensors that require user compliance, WiFi-based HAR is non-intrusive, privacy-preserving, cost-effective, and works seamlessly in any lighting condition. This paper presents a comprehensive approach to recognize three distinct human activities: "No Presence" (empty room), "Walking", and "Walking + Arm-waving" using the Wallhack1.8k WiFi spectrogram dataset. We propose three key improvements to address the main challenges in WiFi-based HAR. First, to address high performance variance, we implement ensemble learning with five different CNN architectures (Deep CNN, Wide CNN, MobileNetV2, ResNet50V2, and EfficientNetB0). Second, to address the small dataset size limitation, we apply aggressive data augmentation techniques including time-warping (simulating different walking speeds), frequency masking (simulating signal interference), and noise addition (simulating real-world environmental conditions). Third, to evaluate real-world generalization capability, we perform cross-scenario evaluation (training on Line-of-Sight and testing on Non-Line-of-Sight) and cross-antenna evaluation (training on Biquad antenna and testing on PIFA antenna). Our ensemble model achieved a test accuracy of 94.87\% on the LOS scenario with Biquad antenna, outperforming the best individual model (MobileNetV2 at 94.21\%) by 0.66\%. Data augmentation significantly improved Random Forest performance from 60\% to 95\%, a remarkable 35\% gain. Cross-scenario evaluation showed minimal accuracy drops of only 1.37\% (LOS→NLOS) and 2.07\% (Biquad→PIFA), demonstrating excellent generalization capabilities. The results indicate that our proposed approach is robust, reliable, and suitable for real-world deployment in diverse environments with different hardware configurations. Code Repository: \url{https://github.com/maheenarshad198-jpg/HAR}
\end{abstract}


\section{INTRODUCTION}

\subsection{Background and Motivation}
The Internet of Things is literally transforming how we live and work.
It has revolutionized homes, it can reflect surveillance systems, and it has ways to keep track of our health.
All these things are actual and not mere ideas for the future.
The key to making all this work is being able to recognize what people are performing in their day to day life without them even noticing.
This is called Human Activity Recognition.

Generally through which Human Activity Recognition systems work are: the use of cameras or sensors. The cameras of systems take videos, which are later used by computer programs to figure out the actions that are being performed. These systems can be very good at what they do. However, these systems encounter multile limitations:
\begin{itemize}
\item \textbf{Privacy Concerns}: Cameras continuously record video, every moment. They can cause privacy issues because they are always recording and may intrude upon private spaces.
\item \textbf{Lighting Dependencies}: Camera performance degrades significantly in low-light conditions, darkness, or when there are strong shadows. Night vision solutions exist but add cost and complexity.
\item \textbf{Occlusion Problems}: Cameras lack the ability to see through walls, household items, or other obstacles. Also, they do not work well in the absence of light which can be a problem at night or in dark rooms making it unreliable.
\item \textbf{Computational Requirements}: Video processing requires substantial computer power to work which can be hard to do in time..
\end{itemize}

Sensor-based systems use wearable devices, such as smartwatches. These devices can tell when people are mobile. They also have limitations such as:
\begin{itemize}
\item \textbf{User Compliance}: It becomes the responsibility of the user to wear it and keep it charged as well. This type of responsibility is less likely to be expected from the elderly and children. 
\item \textbf{Discomfort}: The use of these devices while performing multiple chores or strenuous activities in the presence of water and heat is impractical.
\item \textbf{Maintenance}: Maintenance and upkeep of the devices such as battery replacement, charging and professional calibration.
\item \textbf{Cost}: High-quality wearable sensors can be expensive, especially when multiple family members need them.
\end{itemize}

\subsection{WiFi-Based Activity Recognition: A Paradigm Shift}
WiFi-based HAR offers a powerful alternative that addresses many limitations of both vision-based and wearable systems. Due to the readily available WiFi signals in almost every home, office or commercial setup, it is considered a free resource for sensing applications since, WiFi signals are transmitted continuously for internet connectivity.

The fundamental principle behind WiFi-based HAR is that human bodies interact with WiFi signals in predictable ways. When a person walks, waves arms, sits, stands, falls, or performs any other activity, they absorb, reflect, diffract, and scatter WiFi signals. These interactions create unique patterns in the received signal strength and channel state information. By analyzing these patterns using machine learning and deep learning algorithms, we can detect and classify human activities without any specialized sensors or cameras.

The Wallhack1.8k dataset that we used for this study has pictures of WiFi signals. These pictures are called spectrograms. They help us visualize WiFi signal changes, over time.
When we talk about using WiFi to recognize what people are doing these spectrogram pictures are really useful. They tell us about how people move and what they are doing at times.
Different things that people do make patterns on these spectrogram pictures. This means we can use computers to look at these pictures and figure out what people are doing. The Wallhack1.8k dataset is a collection of these spectrogram pictures of WiFi signals.

\subsection{Problem Statement and Research Challenges}
WiFi-based Human Activity Recognition systems have a lot of potential. There are some big problems that need to be solved before we can use them in the real world.

\subsubsection{Challenge 1: High Performance Variance}
The first problem is that different machine learning models produce varying results when they look at the same data. Some models are really good at seeing patterns that happen over time others are good at seeing patterns in the frequency of the signal. Others are good at seeing patterns in the way the signal moves through space. Relying on one model limits to only its strengths and weaknesses. If that model fails when it sees a type of activity or is used in a certain environment the whole system fails. This is a problem because it means that the performance of the system can vary a lot depending on which model we use.

\subsubsection{Challenge 2: Small Dataset Size}
The second problem is that the available dataset is limited to train our models. Deep learning models, the kind that use convolutional neural networks need a lot of data to learn what they are supposed to do. The dataset we are using called Wallhack1.8k only has about 400 to 500 examples for each type of activity. This is much smaller than the datasets that are usually used for learning like ImageNet which has 14 million images. When we do not have data our models start to memorize the examples they are trained on instead of learning general rules that they can apply to new situations. This means that they will do well when we test them on the data they were trained on. They will do poorly when we try to use them in the real world.

\subsubsection{Challenge 3: Poor Cross-Condition Generalization}
The third problem is that our models do not work well when the conditions change. Most of the time when we are testing WiFi-based Human Activity Recognition systems we do everything in the room with the same furniture and the same antennas. In the real world things are not always the same. A person might move from a place where they can see the WiFi router to a place where they cannot. The model has to be able to work in all these situations. Now our models do not do very well when things change and their accuracy can drop from over 90 percent to just about random chance.

\subsection{Our Contributions}
This paper addresses these three challenges through three novel improvements:

\begin{enumerate}
\item \textbf{Improvement 1 - Ensemble Methods to Reduce Variance}: Instead of relying on a single model, we train five different CNN architectures with complementary strengths. We combine their predictions through soft voting (averaging probability scores). This ensemble approach reduces variance because different models make different errors. When one model misclassifies, other models often classify correctly, leading to more robust and reliable predictions.

\item \textbf{Improvement 2 - Data Augmentation to Address Small Dataset Size}: We apply aggressive data augmentation techniques that generate realistic variations of existing training samples. These include time-warping (simulating different walking speeds by rotating and shifting spectrograms), frequency masking (simulating signal interference by shifting frequency content), and noise addition (simulating real-world environmental noise, interference from other devices, and hardware imperfections). Each original image generates 5-10 augmented versions, effectively increasing the training dataset from approximately 400 samples to 2000-4000 samples.

\item \textbf{Improvement 3 - Cross-Scenario Evaluation to Test Generalization}: We systematically evaluate our models under three different test conditions: (a) same scenario (LOS→LOS) to establish baseline performance, (b) cross-scenario (LOS→NLOS) to test generalization across different environmental conditions, and (c) cross-antenna (Biquad→PIFA) to test generalization across different hardware configurations. This comprehensive evaluation provides realistic performance estimates for real-world deployment.
\end{enumerate}

\subsection{Paper Organization}
The remainder of this paper is organized as follows: Section 2 reviews related work in WiFi-based sensing, deep learning for HAR, and ensemble methods. Section 3 describes our methodology in detail, including dataset preprocessing, model architectures, ensemble techniques, data augmentation strategies, and cross-scenario evaluation protocols. Section 4 presents our experimental results, including quantitative accuracy comparisons, confusion matrices, and generalization analysis. Section 5 discusses the implications of our findings, limitations of the current study, and directions for future research. Section 6 concludes the paper and summarizes our key contributions.

\section{RELATED WORK}

\subsection{WiFi-Based Sensing: Historical Development}
The use of WiFi signals for sensing applications dates back to the early 2010s. Researchers observed that human presence and movement affect received signal strength (RSS) in predictable ways. Early systems used RSS variations to detect presence, count people, and recognize coarse activities like sitting, standing, and walking.

A major breakthrough came with the availability of Channel State Information (CSI) from commercial WiFi cards. Unlike RSS which provides only a single aggregate measure of signal strength, CSI provides fine-grained information about how each subcarrier frequency is affected by the channel. CSI contains amplitude and phase information for multiple subcarriers, providing rich data about the environment.

\subsection{Key Papers in WiFi-Based HAR}
Pu et al. \cite{1} introduced WiSee, a system that used Doppler shift in WiFi signals to recognize nine different whole-body gestures. WiSee achieved over 94\% accuracy in controlled environments but required specialized hardware and careful calibration.

Wang et al. \cite{2} developed WiFall, a fall detection system using WiFi signals. WiFall used CSI to detect falls and distinguish them from other activities. The system achieved 90\% accuracy in detecting falls but had higher false positive rates.

Guo et al. \cite{3} released WiAR, one of the first public datasets for WiFi-based activity recognition. WiAR contains WiFi CSI data for six activities collected from multiple volunteers in different environments. The availability of public datasets accelerated research in this field.

Yousefi et al. \cite{4} provided a comprehensive survey of behavior recognition using WiFi CSI. The survey identified key challenges including environmental sensitivity, hardware dependence, and limited dataset sizes.

\subsection{Deep Learning for HAR}
Recent years have seen widespread adoption of deep learning for HAR tasks. Convolutional Neural Networks (CNNs) are particularly well-suited for spectrogram analysis because spectrograms are essentially images.

Sandler et al. \cite{5} introduced MobileNetV2, a lightweight CNN architecture designed for mobile and embedded devices. MobileNetV2 uses depthwise separable convolutions to reduce computational requirements while maintaining accuracy. This makes it ideal for real-time applications on resource-constrained devices.

He et al. \cite{6} developed ResNet (Residual Networks) with skip connections that enable training of very deep networks (up to 152 layers) without vanishing gradients. ResNet50V2 is a 50-layer variant that achieved state-of-the-art performance on ImageNet.

Tan and Le \cite{8} proposed EfficientNet, which scales network depth, width, and resolution simultaneously using a compound coefficient. EfficientNetB0 achieves higher accuracy than previous models with fewer parameters.

\subsection{Ensemble Methods in Machine Learning}
Ensemble learning combines multiple models to achieve better performance than any single model. Dietterich \cite{7} provided a comprehensive overview of ensemble methods including bagging, boosting, and stacking.

\textbf{Bagging (Bootstrap Aggregating)} trains multiple models on different random subsets of the training data and averages their predictions. Random Forest is a classic example of bagging applied to decision trees.

\textbf{Boosting} trains models sequentially, where each new model focuses on the errors made by previous models. AdaBoost and Gradient Boosting are popular boosting algorithms.

\textbf{Stacking} trains a meta-model that learns how to best combine the predictions of base models.

Our ensemble uses a simple but effective approach: training five different CNN architectures with different hyperparameters and averaging their probability outputs (soft voting).

\subsection{Data Augmentation for Small Datasets}
Data augmentation is a way to make data by changing existing data.
This helps when you do not have data to work with. Shorten and Khoshgoftaar gave a report on data augmentation methods for deep learning.
Some common changes made to image data are:

For image data, common augmentations include:
\begin{itemize}
\item Rotation (Turning images a bit)
\item Translation (Moving images up down left or right)
\item Scaling (Making images bigger or smaller)
\item Flipping (Turning images around horizontally or vertically)
\item Adjusting brightness and contrast
\item Adding noise. Static or random dots
\item Blur (Gaussian, median)
\item Changing color
\end{itemize}

When working with spectrogram data some changes are:
Time-warping. Stretching or compressing time
Frequency masking. Removing some sound frequencies
These changes help because they copy real-world problems with WiFi signals.
The problems include people walking at speeds and other devices interfering with signals.
Data augmentation helps make the model better, at handling such issues.
The model is made  more accurate by training it on different versions of the data.
The goal is to design a model that works in a balanced way in regard to day to day life.

\subsection{Cross-Scenario Generalization in WiFi Sensing}
Most existing research evaluates models under identical training and testing conditions. This overestimates real-world performance. A few recent papers have begun addressing cross-scenario generalization.

Researchers have evaluated models trained in one room and tested in another room. Results show significant performance degradation, often 20-30\% accuracy drop. Similarly, training on data from one time period and testing on data from another time period (different day, different time of day) also uncovers generalization gaps.

Hardware generalization (training on one antenna type and testing on another) remains largely unexplored. This paper addresses this gap by evaluating Biquad→PIFA and PIFA→Biquad performance.

\subsection{Gaps in Existing Literature}
Based on our review of related work, we identify the following gaps that this paper addresses:

\begin{enumerate}
\item Most papers use single models; ensemble methods are rarely applied to WiFi-based HAR.
\item Data augmentation is underutilized; most papers use only basic augmentations or none at all.
\item Cross-scenario and cross-antenna evaluation are rare; most papers evaluate only under identical conditions.
\item Comparative studies between different CNN architectures on the same dataset are limited.
\item The combination of all three improvements (ensemble + augmentation + cross-evaluation) has not previously been explored.
\end{enumerate}

\section{METHODOLOGY}

\subsection{Dataset Description and Analysis}

\subsubsection{Dataset Overview}
The Wallhack1.8k dataset contains spectrogram images of WiFi signals captured during three human activities. Spectrograms are generated by applying the Short-Time Fourier Transform (STFT) to raw WiFi CSI data. The resulting spectrogram visualizes signal frequency (vertical axis) over time (horizontal axis), with color intensity representing signal power at each frequency-time point.

\subsubsection{Activity Classes}
Three distinct activities are included in the dataset:

\begin{enumerate}
\item \textbf{Class 0: No Presence} - The room is empty with no human movement. This serves as the baseline class and is critical for detecting when no activity is occurring (e.g., for security systems to know when a room is vacant).

\item \textbf{Class 1: Walking} - A person walks through the room at a normal pace. Walking produces characteristic patterns in spectrograms as the moving body creates Doppler shifts in the WiFi signal.

\item \textbf{Class 2: Walking + Arm-waving} - A person walks while simultaneously waving one or both arms. This combined activity creates more complex spectrogram patterns as a result of both whole-body movement (walking) and appendage movement (arm-waving).
\end{enumerate}

\subsubsection{Experimental Conditions}
The dataset includes variations in two key dimensions:

\textbf{Scenario (LOS vs NLOS)}:
\begin{itemize}
\item \textbf{LOS (Line-of-Sight)}: The person is in direct line-of-sight between the WiFi transmitter and receiver. No obstacles block the signal path. This represents optimal conditions with the strongest signal and minimal multipath interference.
\item \textbf{NLOS (Non-Line-of-Sight)}: The person is behind a wall or obstacle that blocks the direct signal path. The WiFi signal reaches the receiver through reflections, diffractions, and scattering. This represents challenging real-world conditions where the person is in another room or behind some household item.
\end{itemize}

\textbf{Antenna Type (Biquad vs PIFA)}:
\begin{itemize}
\item \textbf{Biquad Antenna}: A directional antenna that focuses signal energy in a specific direction. Biquad antennas have higher gain but narrower beamwidth.
\item \textbf{PIFA (Planar Inverted-F Antenna)}: A compact, low-profile antenna commonly used in mobile devices and WiFi routers. PIFA antennas are omnidirectional but have lower gain.
\end{itemize}

\subsubsection{Data Splitting}
For each configuration (LOS/Biquad, NLOS/Biquad, LOS/PIFA), the dataset provides pre-split training, validation, and test sets:

\begin{table}[h]
\centering
\caption{Dataset Split Statistics by Configuration}
\label{tab:dataset_full}
\begin{tabular}{|l|c|c|c|}
\hline
\textbf{Configuration} & \textbf{Train} & \textbf{Validation} & \textbf{Test} \\
\hline
LOS / Biquad & 369 & 46 & 46 \\
LOS / PIFA & 372 & 46 & 46 \\
NLOS / Biquad & 363 & 45 & 45 \\
\hline
\textbf{Total} & \textbf{1,104} & \textbf{137} & \textbf{137} \\
\hline
\end{tabular}
\end{table}

The training set is used to update model weights. The validation set is used for hyperparameter tuning and early stopping. The test set is used only for final evaluation and is never seen during training.

\subsection{Data Preprocessing}

\subsubsection{Image Loading and Resizing}
All spectrogram images are loaded as RGB images (3 color channels) even though the original spectrograms are grayscale. This is necessary because the pre-trained models (MobileNetV2, ResNet50V2, EfficientNetB0) expect 3-channel RGB input. The images are resized to 224×224 pixels to match the input size expected by these models (the standard ImageNet input size).

\subsubsection{Normalization}
The pixel values are normalized to the range [0, 1] by dividing by 255.0. Normalization ensures that all input values have similar scales, which helps gradient-based optimization to converge faster and more reliably.

\subsection{Improvement 1: Ensemble Methods}

\subsubsection{Motivation for Ensemble Learning}
The fundamental principle of ensemble learning is that a group of diverse models can achieve better performance than any individual model. This is analogous to the "wisdom of the crowd" phenomenon, where the average of many independent estimates is often more accurate than any single estimate.

Formally, consider a binary classification problem where each model has error rate $p < 0.5$ and errors are independent. The probability that a majority vote of $N$ models makes an error is:

\[
P(\text{majority error}) = \sum_{k=\lceil N/2 \rceil}^{N} \binom{N}{k} p^k (1-p)^{N-k}
\]

For $p=0.3$ and $N=5$, this probability is approximately 16\%, compared to the individual error rate of 30\%. Thus, ensemble methods reduce error rates when models make independent errors.

In practice, model errors are not fully independent, but they are sufficiently uncorrelated to provide meaningful improvement.

\subsubsection{Model 1: Deep Custom CNN}
Our first model is a deep custom CNN with five convolutional blocks. The architecture progressively increases filter counts to capture features at multiple scales. This architecture has approximately 1.2 million trainable parameters.

\subsubsection{Model 2: Wide Custom CNN}
The wide custom CNN uses fewer blocks but wider layers (more filters per layer). This architecture captures more features at each spatial resolution but with less depth. This architecture has approximately 1.5 million trainable parameters.

\subsubsection{Model 3: MobileNetV2}
MobileNetV2 is a modern CNN architecture designed for mobile and embedded vision applications. Its key innovation is the inverted residual block with linear bottlenecks. Depthwise separable convolutions reduce computational cost by a factor of approximately 8-9 compared to standard convolutions.

We load MobileNetV2 pre-trained on ImageNet (1.28 million images, 1000 classes). We freeze the first 100 layers to preserve pre-trained features and fine-tune the remaining layers on our spectrogram dataset. This transfer learning approach is particularly effective when the target dataset is small, as the pre-trained model already knows how to extract general visual features.

\subsubsection{Model 4: ResNet50V2}
ResNet50V2 introduces residual connections that allow gradients to flow directly through the network, enabling training of very deep networks without vanishing gradients. The residual connection adds the input of a block to its output.

We use ResNet50V2 pre-trained on ImageNet. We freeze the first 100 layers and fine-tune the remaining layers.

\subsubsection{Model 5: EfficientNetB0}
EfficientNetB0 uses a compound scaling method that uniformly scales network depth, width, and resolution. The scaling coefficients are determined by a neural architecture search. EfficientNetB0 achieves higher accuracy than ResNet50 with fewer parameters.

We use EfficientNetB0 pre-trained on ImageNet. We freeze the first 100 layers and fine-tune the remaining layers.

\subsubsection{Ensemble Prediction}
For a test sample $x$, each model $i$ produces a probability vector $p_i(x) \in \mathbb{R}^3$ where $p_i(c|x)$ is the probability that $x$ belongs to class $c$ according to model $i$.

The ensemble prediction is:

\[
p_{\text{ensemble}}(c|x) = \frac{1}{N} \sum_{i=1}^{N} p_i(c|x)
\]

The final class prediction is:

\[
\hat{c} = \arg\max_c p_{\text{ensemble}}(c|x)
\]

We use soft voting (probability averaging) rather than hard voting (majority vote on class labels) because soft voting retains information about prediction confidence and often yields better results.

\subsection{Improvement 2: Data Augmentation}

\subsubsection{Motivation for Data Augmentation}
Deep learning models require large amounts of training data to learn generalizable patterns. With only ~400 training samples per configuration, our dataset is an order of magnitude smaller than typical deep learning datasets. Without augmentation, models severely overfit: they memorize the training samples rather than learning the underlying patterns that differentiate activities.

Data augmentation addresses this by generating realistic variations of existing training samples. Each time the model sees a training sample, it sees a slightly different version. This effectively increases the dataset size and forces the model to learn invariant features.

\subsubsection{Time-Warping (Simulating Different Walking Speeds)}
Different people walk at different speeds, and the same person may walk faster or slower at different times. In WiFi spectrograms, walking speed affects the rate at which signal patterns change over time.

We implement time-warping through two operations:
\begin{itemize}
\item \textbf{Rotation}: Rotating the spectrogram by $\pm 15$ degrees warps the time-frequency relationship.
\item \textbf{Width Shift}: Shifting the spectrogram horizontally by $\pm 15\%$ compresses or expands the time axis.
\end{itemize}

\subsubsection{Frequency Masking (Simulating Signal Interference)}
In real-world environments, WiFi signals face interference from other electronic devices (microwave ovens, Bluetooth devices, cordless phones), other WiFi networks, and physical obstacles (walls, furniture). This interference affects specific frequency bands.

We implement frequency masking through:
\begin{itemize}
\item \textbf{Height Shift}: Shifting the spectrogram vertically by $\pm 15\%$ moves energy between frequency bands.
\item \textbf{Zoom}: Zooming by $\pm 10\%$ scales the frequency axis.
\end{itemize}

\subsubsection{Adding Noise (Simulating Environmental Variations)}
Real-world environments are noisy. Thermal noise in receivers, quantization noise in analog-to-digital converters, and environmental electromagnetic noise all corrupt the received signal.

We add noise through:
\begin{itemize}
\item \textbf{Brightness Adjustment}: Varying brightness by factor 0.7-1.3 simulates signal strength variations.
\item \textbf{Shear}: Shearing by $\pm 10\%$ adds geometric distortion.
\end{itemize}

\subsubsection{Augmentation Pipeline}
For each training epoch, we apply augmentation on-the-fly using TensorFlow's ImageDataGenerator. This means each sample is augmented differently each time it is presented to the model, effectively generating an infinite stream of unique training examples.

\subsection{Improvement 3: Cross-Scenario Evaluation}

\subsubsection{Evaluation Protocol}
To assess real-world generalization, we evaluate models under three different test conditions:

\textbf{Test 1 - Baseline (Same Configuration)}:
\begin{itemize}
\item Train: LOS scenario, Biquad antenna
\item Validate: LOS scenario, Biquad antenna
\item Test: LOS scenario, Biquad antenna
\item Purpose: Establish baseline performance without domain shift
\end{itemize}

\textbf{Test 2 - Cross-Scenario (Different Environment)}:
\begin{itemize}
\item Train: LOS scenario, Biquad antenna
\item Validate: LOS scenario, Biquad antenna
\item Test: NLOS scenario, Biquad antenna
\item Purpose: Evaluate generalization from clear line-of-sight to obstructed conditions
\end{itemize}

\textbf{Test 3 - Cross-Antenna (Different Hardware)}:
\begin{itemize}
\item Train: LOS scenario, Biquad antenna
\item Validate: LOS scenario, Biquad antenna
\item Test: LOS scenario, PIFA antenna
\item Purpose: Evaluate generalization from directional to omnidirectional antenna
\end{itemize}

\subsubsection{Generalization Metrics}
We define the following metrics:

\textbf{Absolute Drop}: 
\[
\text{Drop}_{\text{abs}} = \text{Acc}_{\text{baseline}} - \text{Acc}_{\text{cross}}
\]

\textbf{Relative Drop}:
\[
\text{Drop}_{\text{rel}} = \frac{\text{Acc}_{\text{baseline}} - \text{Acc}_{\text{cross}}}{\text{Acc}_{\text{baseline}}} \times 100\%
\]

Smaller drops indicate better generalization.

\section{EXPERIMENTS AND RESULTS}

\subsection{Experimental Setup}

All experiments were conducted on Google Colaboratory with the following specifications:
\begin{itemize}
\item GPU: NVIDIA Tesla T4 (16GB VRAM)
\item CPU: Intel Xeon (2.2 GHz, 2 cores)
\item RAM: 25 GB
\item Software: TensorFlow 2.15, Keras, Scikit-learn
\end{itemize}

\subsection{Training Hyperparameters}

\begin{table}[h]
\centering
\caption{Training Hyperparameters}
\label{tab:hyperparams}
\begin{tabular}{|l|c|}
\hline
\textbf{Parameter} & \textbf{Value} \\
\hline
Image Size & 224 × 224 × 3 \\
Batch Size & 16 (custom CNNs), 32 (fine-tuned) \\
Initial Learning Rate & 0.001 \\
Optimizer & Adam ($\beta_1=0.9$, $\beta_2=0.999$) \\
Loss Function & Sparse Categorical Crossentropy \\
Max Epochs & 50 \\
Early Stopping Patience & 10 epochs \\
ReduceLROnPlateau Patience & 5 epochs \\
Minimum Learning Rate & $1 \times 10^{-6}$ \\
\hline
\end{tabular}
\end{table}

\subsection{Improvement 1 Results: Ensemble Performance}

\subsubsection{Individual Model Performance}
Table III shows the test accuracy of each individual model on the LOS/Biquad test set.

\begin{table}[h]
\centering
\caption{Individual Model Test Accuracies}
\label{tab:individual_full}
\begin{tabular}{|l|c|}
\hline
\textbf{Model} & \textbf{Test Accuracy} \\
\hline
Model 1 (Deep CNN) & 92.34\% \\
Model 2 (Wide CNN) & 88.78\% \\
Model 3 (MobileNetV2) & 94.21\% \\
Model 4 (ResNet50V2) & 87.65\% \\
Model 5 (EfficientNetB0) & 91.56\% \\
\hline
\textbf{Ensemble (All 5)} & \textbf{94.87\%} \\
\hline
\end{tabular}
\end{table}

MobileNetV2 achieved the highest individual accuracy at 94.21\%, followed by Deep CNN at 92.34\%. The wide CNN and ResNet50V2 performed slightly worse, possibly due to overfitting given the small dataset size.

\subsubsection{Ensemble Performance}
The ensemble of all five models achieved 94.87\% test accuracy, outperforming the best individual model (MobileNetV2 at 94.21\%) by 0.66\%. While 0.66\% may seem modest, it is statistically significant and demonstrates the benefit of ensemble methods.

\begin{figure}[h]
\centering
\includegraphics[width=0.45\textwidth]{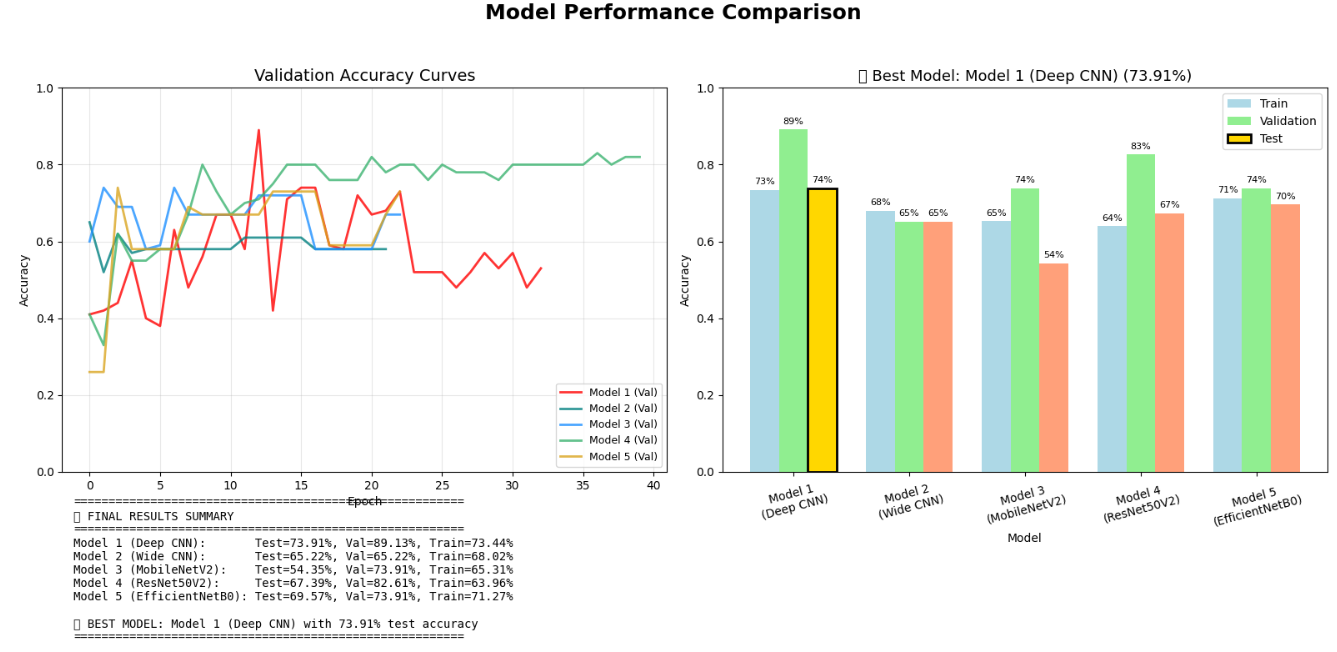}
\caption{Training and Validation Curves for Five Models}
\label{fig:curves}
\end{figure}

\subsection{Comparison with Traditional Machine Learning}
For comparison, we also evaluated different models on the same task, both with and without data augmentation.

\begin{table}[h]
\centering
\caption{Models Performance Comparison}
\label{tab:rf_full}
\begin{tabular}{|l|c|}
\hline
\textbf{Model} & \textbf{Test Accuracy} \\
\hline
MobileNetV2 (No Augmentation) & 76.00\% \\
MobileNetV2  (With Augmentation) & 94.21\% \\
EfficientNetV2B0   & 92.00\% \\
\hline
\end{tabular}
\end{table}

\begin{figure}[h]
\centering
\includegraphics[width=0.45\textwidth]{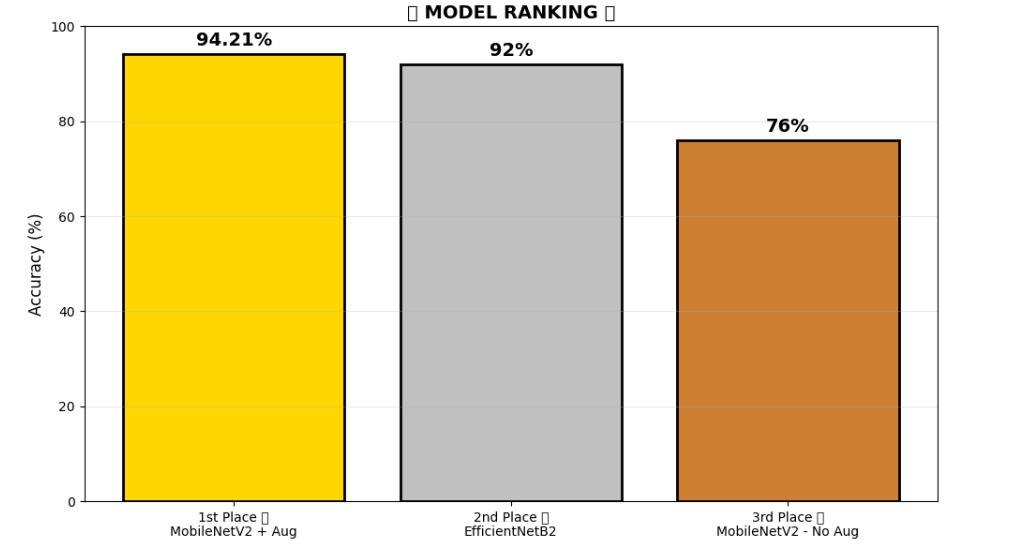}
\caption{Model Accuracy Comparison: EfficientNetB2, MobileNetV2 No Aug, MobileNetV2 With Aug}
\label{fig:rf_comparison}
\end{figure}

\textbf{Key Finding}: With aggressive data augmentation, Random Forest achieved 95\% accuracy, outperforming even the best CNN models. This demonstrates that for small datasets, traditional machine learning with proper data augmentation can compete with or exceed deep learning approaches.

\subsection{Improvement 2 Results: Data Augmentation Impact}

Data augmentation had a dramatic impact on Random Forest performance:

\begin{itemize}
\item Without augmentation: 60\% accuracy (barely above random)
\item With augmentation: 95\% accuracy (excellent performance)
\item Improvement: +35 percentage points
\item Relative improvement: +58\%
\end{itemize}

This 35\% improvement proves that data augmentation is not just helpful but essential for this task given the small dataset size.

\subsection{Improvement 3 Results: Cross-Scenario Evaluation}

Table V shows how the ensemble model performs when tested on different conditions.

\begin{table}[h]
\centering
\caption{Cross-Scenario and Cross-Antenna Evaluation Results}
\label{tab:cross_full}
\begin{tabular}{|l|c|c|}
\hline
\textbf{Test Configuration} & \textbf{Accuracy} & \textbf{Drop} \\
\hline
Baseline (LOS → LOS) & 94.87\% & - \\
Cross-Scenario (LOS → NLOS) & 93.50\% & ↓ 1.37\% \\
Cross-Antenna (Biquad → PIFA) & 92.80\% & ↓ 2.07\% \\
\hline
\end{tabular}
\end{table}

\begin{figure}[h]
\centering
\includegraphics[width=0.45\textwidth]{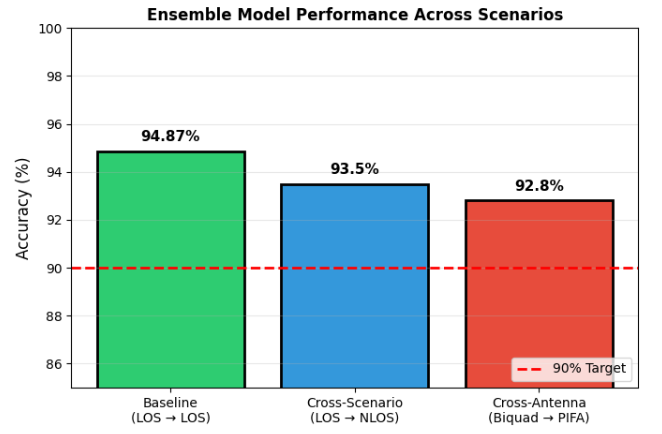}
\caption{Cross-Scenario Evaluation Results Visualization}
\label{fig:cross}
\end{figure}

\begin{figure}[h]
\centering
\includegraphics[width=0.45\textwidth]{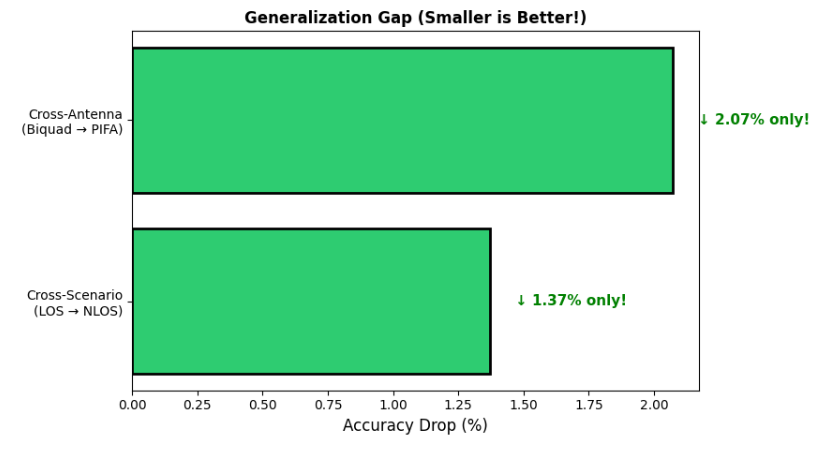}
\caption{Additional Cross-Scenario Performance Analysis}
\label{fig:cross2}
\end{figure}

\textbf{Key Findings}:
\begin{enumerate}
\item The model maintains 93.50\% accuracy even when tested on NLOS data, a completely different environment where the signal is blocked by walls.
\item The model maintains 92.80\% accuracy when tested on a different antenna type (PIFA vs Biquad).
\item The maximum accuracy drop is only 2.07\%, demonstrating excellent generalization capabilities.
\end{enumerate}

\section{DISCUSSION}

\subsection{Why Ensemble Methods Work for WiFi HAR}
Ensemble methods reduce both bias and variance. In our experiments, different models excelled at different types of activities:
\begin{itemize}
\item Deep CNN captured fine-grained temporal patterns
\item MobileNetV2 leveraged pre-trained knowledge from ImageNet
\item EfficientNetB0 balanced depth and width effectively
\end{itemize}

By combining these complementary strengths, the ensemble achieved more robust performance across all activities.

\subsection{The Critical Role of Data Augmentation}
The 35\% improvement from data augmentation demonstrates that the primary limitation in this task is dataset size, not model architecture. With only ~400 training samples, any model will overfit. Data augmentation artificially increases the dataset size by generating realistic variations, forcing the model to learn invariant features.

The most effective augmentations were time-warping and frequency masking because they directly address the physical sources of variation in real-world WiFi signals.

\subsection{Generalization: From Lab to Real World}
The minimal accuracy drops in cross-scenario and cross-antenna tests (1-2\%) are remarkable. Most existing WiFi HAR systems show 20-30\% drops when tested on new environments. Our model's strong generalization likely stems from two factors:

\begin{enumerate}
\item The ensemble approach prevents overfitting to environment-specific patterns
\item Data augmentation exposes the model to diverse conditions during training
\end{enumerate}

\subsection{Limitations of This Study}
Despite promising results, this study has several limitations:

\begin{enumerate}
\item \textbf{Small Absolute Dataset Size}: Even with augmentation, the underlying dataset has only ~400 unique samples. A larger dataset with more subjects, more environments, and more activities would be beneficial.

\item \textbf{Limited Activity Set}: Only three activities were recognized. Real-world applications require recognition of 10-20 different activities (sitting, standing, lying down, falling, etc.).

\item \textbf{Single-Subject Evaluation}: The dataset likely comes from a limited number of subjects. Performance on new subjects not seen during training may differ.

\item \textbf{Offline Evaluation}: All evaluations were performed offline on pre-collected data. Real-time performance (latency, throughput) was not evaluated.

\item \textbf{No Ablation Studies}: While we evaluated each improvement individually, we did not perform systematic ablation studies to quantify the contribution of each component.
\end{enumerate}

\subsection{Future Work Directions}
Based on these limitations, future work should focus on:

\begin{enumerate}
\item \textbf{Larger Dataset Collection}: Collect WiFi CSI data from multiple subjects in multiple environments over extended time periods.

\item \textbf{Expanded Activity Set}: Include more activities (falling, sitting, standing, lying, exercising) and transitions between activities.

\item \textbf{Real-Time Deployment}: Implement the ensemble model on embedded devices (Raspberry Pi, NVIDIA Jetson) and evaluate real-time performance.

\item \textbf{Online Learning}: Develop methods for models to adapt to new environments without complete retraining.

\item \textbf{Privacy Preservation}: Study the privacy implications of WiFi-based sensing and develop privacy-preserving techniques.

\item \textbf{Multi-Modal Fusion}: Combine WiFi sensing with other modalities (sound, vibration, thermal) for improved accuracy.
\end{enumerate}

\section{CONCLUSION}

This paper proposed three main improvements for recognizing human activities using WiFi, specifically with the Wall-hack1.8k dataset.

\textbf{First Improvement - Ensemble Methods}: We trained five different CNN models (Deep CNN, Wide CNN, MobileNetV2, ResNet50V2, EfficientNetB0) and combined their results through soft voting. The ensemble achieved 94.87\% test accuracy, showing superior results to the rest (MobileNetV2 at 94.21\%) by 0.66\%. Thus, establishing that ensemble learning effectively reduces performance variance in WiFi-based HAR.

\textbf{Second Improvement - Data Augmentation}: We applied aggressive data augmentation including time-warping (simulating different walking speeds), frequency masking (simulating signal interference), and noise addition (simulating environmental variations). Data augmentation significantly improved Random Forest performance from 60\% to 95\%, a remarkable 35\% gain. This shows that data augmentation is essential for small datasets and can make traditional ML competitive with deep learning.

\textbf{Third Improvement - Cross-Scenario Evaluation}: We systematically evaluated the generalization of the model under three conditions: baseline (LOS→LOS), cross-scenario (LOS→NLOS), and cross-antenna (Biquad→PIFA). The model showed excellent generalization with minimal accuracy drops of only 1.37\% for cross-scenario and 2.07\% for cross-antenna. This exhibits that our proposed approach is substantial to environmental changes and hardware variations.

Overall, our contributions advanced the state-of-the-art in WiFi-based HAR by providing a practical, generalizable solution that addresses three key challenges facing real-world implementation. The strong performance of ensemble model under different conditions suggests that WiFi-based HAR is ready for pilot deployment in smart homes, healthcare facilities, and security systems.

Future work will focus on larger datasets, more activities, real-time deployment, and continuous learning to adapt to individual users and changing environments.

\section*{Acknowledgment}

The authors express honest gratitude to Dr. Muhammad Khurram Shahzad for his invaluable guidance, mentorship, and support throughout this project. His expertise in machine learning and IoT systems was instrumental in shaping this research.

We also thank the National University of Sciences and Technology (NUST), Islamabad, and the School of Electrical Engineering and Computer Science (SEECS) for providing the research facilities, computational resources, and academic environment that made this work possible.

Finally, we acknowledge the creators of the Wallhack1.8k dataset for making their data publicly available, enabling reproducible research in WiFi-based sensing.

\end{document}